\documentclass[letterpaper]{article}
\usepackage{uai2020}
\usepackage[margin=1in]{geometry}

\usepackage{times}

\usepackage[hyphens]{url}
\usepackage{amssymb}
\usepackage{amsmath}
\usepackage{amsthm}
\usepackage{algorithm}
\usepackage{algpseudocode}
\usepackage{algcompatible}
\usepackage{tikz}
\usepackage{booktabs}
\usepackage[sort]{natbib}
\usepackage{multicol}
\usepackage{mathtools}

\newtheorem{lemma}{Lemma}
\newtheorem{proposition}{Proposition}

\usetikzlibrary{shapes,backgrounds,calc}
\usetikzlibrary{arrows, automata}

\makeatletter
\tikzset{circle split part fill/.style  args={#1,#2}{%
 alias=tmp@name, 
  postaction={%
    insert path={
     \pgfextra{%
     \pgfpointdiff{\pgfpointanchor{\pgf@node@name}{center}}%
                  {\pgfpointanchor{\pgf@node@name}{east}}%
     \pgfmathsetmacro\insiderad{\pgf@x}
      \fill[#1] (\pgf@node@name.base) ([xshift=-\pgflinewidth]\pgf@node@name.east) arc
                          (0:180:\insiderad-\pgflinewidth)--cycle;
      \fill[#2] (\pgf@node@name.base) ([xshift=\pgflinewidth]\pgf@node@name.west)  arc
                           (180:360:\insiderad-\pgflinewidth)--cycle;            
         }}}}}  
 \makeatother

\usepackage{graphicx}
\usepackage{caption}
\usepackage{subcaption}
\usepackage{xcolor}

\usepackage{bm}

\def\K{\mathcal{K}}
\def\y{\mathbf{y}}
\def\z{\mathbf{z}}

  \def\E{\mathbb{E}}

  \def\w{\mathbf{w}}
  \def\W{\mathbf{W}}
  
  \def\1{{\bf{1}}}

  \def\o{\mathbf{o}}
  \def\v{\mathbf{v}}
  \def\O{\mathbf{O}}

  \def\X{\mathcal{X}}

  \def\hr{\hat{r}}

\DeclareMathOperator*{\argmax}{arg\,max}

\newcommand{\WOPT}{W_{\mathit{OPT}}}

\def\submit{0}

\ifx\submit\undefined

\else

\fi

\title{Kidney Exchange with Inhomogeneous Edge Existence Uncertainty}

\author{ 
{\bf Hoda Bidkhori}\\
University of Pittsburgh\\
\And
{\bf John P Dickerson} \\ 
University of Maryland\\
\And 
{\bf Duncan C McElfresh}   \\
University of Maryland\\
\And
{\bf Ke Ren}   \\
University of Pittsburgh\\
}

\begin{document}

\maketitle
\begin{abstract}

Patients with end-stage renal failure often find kidney donors who are willing to donate a life-saving kidney, but who are medically incompatible with the patients.  
Kidney exchanges are organized barter markets that allow such incompatible patient-donor pairs to enter as a single agent---where the patient is endowed with a donor ``item''---and engage in trade with other similar agents, such that all agents ``give'' a donor organ if and only if they receive an organ in return. 
In practice, organized trades occur in large cyclic or chain-like structures, with multiple agents participating in the exchange event.  
Planned trades can fail for a variety of reasons, such as unforeseen logistical challenges, or changes in patient or donor health.
These failures cause major inefficiency in fielded exchanges, as if even one individual trade fails in a planned cycle or chain, \emph{all or most of the resulting cycle or chain fails}.  
%
%
Ad-hoc, as well as optimization-based methods, have been developed to handle failure uncertainty; nevertheless, the majority of the existing methods use very simplified assumptions about failure uncertainty and/or are not scalable for real-world kidney exchanges.

Motivated by kidney exchange, we study a stochastic cycle and chain packing problem, where we aim to identify structures in a directed graph to maximize the expectation of matched edge weights. All edges are subject to failure, and the failures can have nonidentical probabilities. To the best of our knowledge, the state-of-the-art approaches are only tractable when failure probabilities are identical. We formulate a relevant non-convex optimization problem and propose a tractable mixed-integer linear programming reformulation to solve it. In addition, we propose a model that integrates both risks and the expected utilities of the matching by incorporating conditional value at risk (CVaR) into the objective function, providing a robust formulation for this problem. Subsequently, we propose a sample-average-approximation (SAA) based approach to solve this problem. We test our approaches on data from the United Network for Organ Sharing (UNOS) and compare against state-of-the-art approaches.  Our model provides better performance with the same running time as a leading deterministic approach (PICEF). Our CVaR extensions with an SAA-based method improves the $\alpha \times 100\%$ ($0<\alpha\leqslant 1$) worst-case performance substantially compared to existing models.

\end{abstract}

\section{INTRODUCTION}\label{sec:intro}

Kidney exchange is a centralized barter market were patients with end-stage renal disease trade willing donors in cyclic or chain-like transactions~\citep{Rapaport86:Case,Roth04:Kidney,Abraham07:Clearing}.
The aim of the kidney exchange clearinghouse is to find the ``best'' disjoint set of such swaps---i.e., to solve a cycle and chain packing problem.  
Exchanges already account for over 12\% of living kidney donations in the US, and exchange programs are growing worldwide~\citep{Biro19:Building}---including via extensions to liver and lung~\citep{Ergin17:Multi}, and even multi-organ~\citep{Dickerson17:MultiJAIR}, exchange.  
Fielded exchanges face several source of inefficiency, primarily due to pre-transplant ``failure''~\citep{Leishman19:Challenges}; that is, most \emph{planned} transplants never result in transplantation due to medical or logistical incompatability~\citep{Dickerson18:Failure,Glorie14:Kidney,Anderson15:Finding,Alvelos15:Compact,Dickerson16:Position,Goldberg19:Maximum,Manlove15:Paired,Klimentova16:Maximising,Agarwal19:Market}.
In other words, the exchange program cannot be certain whether a compatible patient and donor will result in a transplant.
Exchanges are often represented by directed graphs (see~\S~\ref{sec:model}), where edges indicate potential transplants and edge weights reflect the medical or social utility of the transplant.
If a planned transplant (i.e., \emph{edge}) fails, its effects can cascade through the exchange, causing other edges to fail (see~\S~\ref{sec:example})---thus, edge failures can severely impact the overall utility of an exchange.
Thus it is of interest for exchange coordinators to account for uncertainty when planning transplants.
Kidney exchange with \emph{edge existence uncertainty}---specifically, selecting transplants which minimize impact from edge failures---is the focus of this paper.

Many approaches have been proposed to deal with edge existence uncertainty, often using stochastic or robust optimization techniques.
However most of these models either make the simplifying assumption that all transplants are equally likely to fail, or they require intractable algorithms that cannot be used on large exchanges.

We fill a major gap in prior work by proposing the first \emph{scalable} algorithm (meaning it uses a number of variables polynomial in the input size) for maximizing expected matching weight, with \emph{non-identical} failure probabilities.
This is an important step forward, as failure probabilities are known to be inhomogeneous--some edges are inherently riskier than others~\citep{Dickerson18:Failure}.
We provide a mixed-integer linear program for our approach, which is compact and can be solved directly by a general-purpose integer programming solver (e.g., CPLEX, Gurobi, or SCIP). 
In computational experiments we demonstrate that accounting for inhomogeneous edge probabilities improves over state-of-the-art approaches, using data from the United Network for Organ Sharing. 

Additionally, we propose a modified version of the kidney exchange problem which balances the \emph{mean expected weight} with the \emph{worst-case} weight (``risk'') of an exchange with known nonidentical edge failure probabilities; we achieve this balance using a conditional value-at-risk (CVaR) objective. 
This approach is motivated by the fact that expected weight can be misleading when the worst-case outcome can be arbitrarily bad (see \S~\ref{sec:example}). 
We are not the first to propose a CVaR approach for kidney exchange; however, previous CVaR-based approaches do not allow for arbitrary length limits on cycles and chains--which are used by all fielded exchanges. 
With cycle and chain length limits, the kidney exchange problem with a CVaR objective is challenging, as there is no closed-form expression for the objective function.
Thus, we propose a sample-average-approximation-based method and develop an equivalent mixed-integer linear programming representation.
Computational experiments show that our model improves the worst-case mean over state-of-the-art methods.

\subsection{Uncertainty in Kidney Exchange}\label{sec:prior}

Many prior approaches address edge existence uncertainty in kidney exchange, often with the objective of maximizing \emph{expected} matching weight, assuming all edges have identical failure probability.
\citet{Dickerson16:Position} provides a scalable formulation in this case, and \citet{Dickerson18:Failure} extends this to consider inhomogeneous edge probabilities; however the latter model can require enumeration of all feasible cycles and chains, which can be intractable for even small exchanges.
Similar approaches have been proposed, but still assume that all edges have equal failure probability~\citep{Alvelos15:Compact,Constantino13:New}.
Rather than maximizing expected edge weight, other approaches take the \emph{risk-averse} perspective, aiming to maximize the worst-case matching weight~\citep{McElfresh19:Scalable,Carvalho20:Robust}; these approaches are often \emph{too} conservative, as the worst case in kidney exchange is often arbitrarily bad (i.e., in the worst case, all planned transplants fail).
\citet{Zheng15:Loss-constrained} propose a CVaR method that endogenously balances structure length with risk; however, their model is not amenable to length caps on cycles and/or chains, a requirement in all fielded kidney exchanges.

Several other optimization-based approaches have been proposed, using recourse~\citep{Anderson15:Finding}, forms of ``fallback'' options~\citep{Manlove15:Paired,Bartier19:Recourse,Wang19:Efficient}, and pre-match edge queries~\citep{Blum13:Harnessing,Blum15:Ignorance,mcelfresh2020Improving}. 
These methods involve additional decision stages, and are not directly comparable in our setting.




Next we describe the formal model of kidney exchange and edge existence uncertainty.

\section{PRELIMINARIES}\label{sec:model}

We represent a kidney exchange as a directed graph $G=(E,V)$ where each vertex $v_i\in V$ is an incompatible patient-donor pair, or a non-directed donor (NDD, i.e., a donor without a paired patient).
Directed edges $e = (v_i,v_j)$ represent potential transplants from the donor of node $i$ to the patient of node $j$; edge weights $w_e>0$ represent the medical or social utility of each potential transplant.
We assume that edge failure probabilities $p_e \in [0,1]$ are known in advance and are not necessarily homogeneous.
That is, if edge $e=(v_i, v_j)$ is matched, then with probability $p_e$ the patient of $v_j$ would still fail to receive a kidney from $v_i$’s donor.

Kidney exchanges consist of two types of swaps: \emph{cycles} consist of several patient donor pairs, while \emph{chains} begin with an NDD and continue through one or more patient pairs~\citep{Roth05:Kidney}.
The goal of the \emph{kidney exchange clearing problem} (KEP) is often to select the set of vertex-disjoint cycles and chains in $G$ which maximize overall edge weight.
We refer to any set of vertex-disjoint cycles and chains as a \emph{matching}.
For example, let $\bm w$ denote the vector of weights for all cycles and chains in the graph, let $\bm x$ denote a vector of binary decision variables, and let $\mathcal M$ denote the set of feasible matchings (i.e., binary vectors $\bm x$ corresponding to vertex-disjoint cycles and chains); in this case the KEP is expressible as $\max_{\bm x\in \mathcal M} \bm x^\top \bm w$.

Cycles and chains are quite vulnerable to edge failure: if \emph{any} edge in a cycle fails, then \emph{none} of the transplants in the cycle can proceed, because at least one of the patients will be left without a compatible donor.
If an edge participating in a chain fails, then none of the edges \emph{following} that failed edge can proceed.\footnote{We assume that chains can be \emph{partially} executed. Some fielded exchanges cancel the entire chain if even one edge fails.}

We consider modified versions of the KEP which account for edge failures, using known edge failure probabilities.
Before describing our approach, we emphasize that the choice of \emph{objective} is important in the KEP.
We demonstrate this point with a small example.

\subsection{Example: Edge Existence Uncertainty}\label{sec:example}

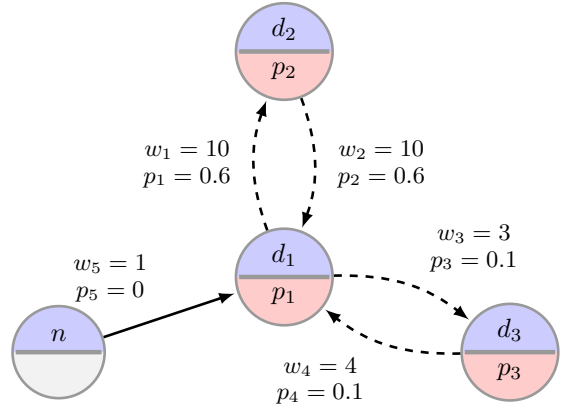
\begin{figure}
\centering
\begin{tikzpicture}[
            > = latex, 
            shorten > = 1pt, 
            auto,
            node distance = 3cm, 
            semithick 
        ]

  \tikzstyle{fake}=[rectangle,minimum size=8mm,opacity=0.0]
  
  \node (alt) [shape=circle split,
    draw=gray!80,
    line width=0.6mm,text=black,font=\bfseries,
    circle split part fill={blue!20,gray!10},
    minimum size=1.2cm,
  ] at (0,0) {$n$\nodepart{lower}};

  \node (p_1) [shape=circle split,
    draw=gray!80,
    line width=0.6mm,text=black,font=\bfseries,
    circle split part fill={blue!20,red!20},
    minimum size=1.2cm,
  ] at (3,1) {$d_1$\nodepart{lower}$p_1$};

    \node (p_3) [shape=circle split,
    draw=gray!80,
    line width=0.6mm,text=black,font=\bfseries,
    circle split part fill={blue!20,red!20},
    minimum size=1.2cm,
  ] at (6, 0) {$d_3$\nodepart{lower}$p_3$};
  
     \node (p_2) [shape=circle split,
    draw=gray!80,
    line width=0.6mm,text=black,font=\bfseries,
    circle split part fill={blue!20,red!20},
    minimum size=1.2cm,
  ] at (3,4) {$d_2$\nodepart{lower}$p_2$};
  
  
%
  

  \path[->] (alt) edge node {{\small $\begin{array}{c}w_5=1 \\ p_5=0\end{array}$}} (p_1);

  \path[dashed,->] (p_1) edge[bend left=20] node {{\small $\begin{array}{c}w_1=10 \\ p_1=0.6\end{array}$}} (p_2);
  \path[dashed,->] (p_2) edge[bend left=20] node {{\small$\begin{array}{c}w_2=10 \\ p_2=0.6\end{array}$}} (p_1);
  \path[dashed,->] (p_1) edge[bend left=20] node {{\small$\begin{array}{c}w_3=3 \\ p_3=0.1\end{array}$}} (p_3);
  \path[dashed,->] (p_3) edge[bend left=20] node {{\small$\begin{array}{c}w_4=4 \\ p_4=0.1\end{array}$}} (p_1);

\end{tikzpicture} 
\caption{Example exchange graph with a single NDD $n$, and three patient-donor pairs; weights $w$ and failure probabilities $p$ are shown for each edge. The max-weight matching is the cycle between pairs $1$ and $2$; the max-expected-weight matching is the cycle between pairs $1$ and $3$, and the risk-averse/robust optimal matching is any the chain beginning with the edge from $n$ to pair $1$.}\label{fig:example-graph}
\end{figure}

The choice of the \emph{objective function}---and, in particular, its treatment of uncertainty---can substantially impact the structure of the final matching.
Consider the exchange in Figure~\ref{fig:example-graph}, in which there are four possible matchings: 2-cycle ($1,2$), 2-cycle $(1,3)$, 3-chain $(n,1,2)$, and 3-chain $(n,1,3)$.\footnote{The 2-chain $(n,1)$ is also a feasible exchange, though this chain has strictly lower weight than either of the 3-chains.}
All edges have integer weight $w$ and failure probability $p$; only the edge from $n$ to pair $1$ is guaranteed to succeed ($p=0$).
\emph{Any} of the four feasible matching in this graph might be ``optimal,'' depending on the choice of objective.

An objective that maximizes overall matching weight (i.e., the objective used by many fielded exchanges~\citep{Anderson15:Finding,UNOS}) would select 2-cycle $(1,2)$ with total weight $10$. 
However this matching is likely to fail: at least one cycle edge will fail with probability $0.84$---in which case the matching receives zero weight. 
Instead, we might maximize \emph{expected} matching weight (e.g., as in~\citet{Dickerson18:Failure}), and select 2-cycle $(1,3)$.
Indeed this matching achieves total expected weight $6.23$, nearly twice the expected weight of cycle $(1,2)$.
Of course, cycle $(1,3)$ has a significant ($\sim 10\%$) chance of failure, which may be unacceptable in a real setting.
Thus, we might choose an objective that aims to maximize the matching weight under the \emph{worst-case} outcome (e.g., as in~\citet{McElfresh19:Scalable}).
In this case, any chain beginning with edge $(n,1)$ is optimal.

Next we describe our approach, beginning with a characterization of the \emph{expected} matching weight.

\tikzstyle{altruist}=[circle,
  thick,
  minimum size=1.0cm,
  draw=black!50!green!80,
  fill=black!20!green!20
]

\tikzstyle{every path}=[line width=1pt]

\begin{table*}\centering
\caption{Comparison of stochastic and robust approaches to kidney exchange, which use a setting comparable to ours.\footnote{There are several other approaches in the literature which allow unlimited cycle and chain length. We do not consider these approaches, because in this case the kidney exchange clearing problem reduces to bipartite matching.} Columns indicate the type of uncertainty considered in the problem (``Unct.'': stochastic or worst-case/robust), whether or not edge failure probability is assumed to be \emph{homogeneous} (``Homog.''), and the number of variables and constraints in each formulation.}\label{tab:method}

\begin{tabular}{@{}ccccc@{}}\toprule
Formulation & Unct. & Homog. & \# Vars. & \# Constr.  \\ \midrule
 PC-TSP \citep{Anderson15:Finding}  & None& N/A & $O(|E|\cdot |V|+ |V|^2 + |C|)$ &  $O(|V|\cdot(|E| + 2^{|V|} + |C|)$  \\
  PICEF \citep{Dickerson16:Position}  & Stoch.& Yes &  $O(L\cdot |E|+ |C|)$ & $O(L\cdot|V|+L\cdot|E|+|C|)$ \\
 ROBUST \citep{McElfresh19:Scalable} & Robust& N/A &  $O(|E|\cdot |V|+ |V|^2 + |C|)$ & $O(|E|\cdot |V|+ |V|^2 + |C|)$ \\
 DPS-18 \citep{Dickerson18:Failure} & Stoch. &  No & $O(|V|^L+|C|)$ & $O(|V|)$ \\
 Our model (\ref{linear:closed2}) & Stoch.& No  & $O(L\cdot |E|+ |C|)$ & $O(L\cdot|V|+L\cdot|E|+|C|)$ \\
\bottomrule
\end{tabular}
\end{table*}

\section{MAXIMIZING EXPECTED MATCHING WEIGHT}\label{sec:max-expected}

We are primarily interested in maximizing the \emph{expected} weight of a matching; indeed this is the focus of most prior work (see~\S~\ref{sec:prior}). 
We refer to this as the \emph{stochastic} KEP.
First we characterize the objective of this problem---the expected matching weight.
With known edge failure probabilities, the expected weight of a cycle or chain is expressible in closed form.

\noindent{\bf Discounted weight of a cycle.}
 The discounted weight of a $k$-cycle $c$ reflects the fact that the \emph{whole cycle} will fail if any single transplant fails. We use $w_e$ to denote the weight of edge $e$ in the cycle, $c$.
 \begin{align*}
     u(c) = \left(\sum_{e \in c}w_e\right)\left[\prod_{e\in c}(1-p_e)\right].
 \end{align*}
\noindent{\bf Discounted weight of a chain.}
The expected weight $u(\kappa)$ of the $k$-chain $\kappa \equiv (v_1,...,v_{k+1})$, where $v_1$ is a non-directed donor (NDD), is defined as
\begin{equation}\label{eq:exp-chains}
     \begin{aligned}
     u(\kappa) = &\sum_{i=2}^{k}p_i \left(\sum_{j=1}^{i-1}w_j\right)\prod_{j=1}^{i-1} (1-p_j) + \\
     &\left(\sum_{i=1}^{k}w_i\right) \prod_{i=1}^{k} (1-p_i).
 \end{aligned}
\end{equation}

In the above, $p_i$ and $w_i$ denotes the failure probability and weight of edge $(v_{i},v_{i+1})$, respectively. The first term above is the sum of expected weights for the chain executing exactly $i-1 = \{1,...,k-1\}$ steps and then failing on the $i$th step. The second term is the resulting weight if the chain executes completely.

Using the above expressions, we can write the stochastic KEP as follows.
With some abuse of notation, let $(\bm C, \bm K)\in \mathcal M$ denote a feasible matching consisting of cycles $\bm C$ and chains $\bm K$.
Problem~\ref{eq:stochastic-kep} is an equivalent formulation of the stochastic KEP.
\begin{equation}\label{eq:stochastic-kep}
    \max_{(\bm C', \bm K') \in \mathcal M} \quad \sum_{c' \in \bm C} u(c' ) + \sum_{\kappa' \in  \bm K} u(\kappa')
\end{equation}
Next we describe our solution approach for Problem~\ref{eq:stochastic-kep}, and an equivalent compact mixed-integer linear program formulation.

\subsection{Compact Formulation for Maximizing Expected Matching Weight}\label{sec:main_model}

Here we present a new compact formulation to maximize the expected weight in the case of \emph{non-identical} edge failure probabilities. \textit{Compact} means that the counts of variables and constraints are polynomial in the size of the input. We compare the size of this model with other state-of-the-art approaches in \S~\ref{sec:main:tractability}.

In~\citep{Dickerson18:Failure}, the authors propose a solution approach for Problem~\ref{eq:stochastic-kep}, which enumerates all feasible cycles and chains in the graph.
However the number of cycles and chains grows exponentially with the size of the graph, meaning this formulation is not compact.
Further, it is intractable to even write this model in memory for large exchanges or long chain lengths.


We propose an exact, compact representation for Problem~\ref{eq:stochastic-kep}, using an equivalent expression for expected chain weight $u(\kappa)$ given in Lemma~\ref{utility:chain}.
\begin{lemma}\label{utility:chain}
The discounted weight $u(\kappa)$ of the $k$-chain $\kappa = (v_1,...,v_{k+1})$ is
\begin{align*}
    u(\kappa) = \sum_{i=1}^{k} w_i\prod_{j=1}^{i} \left(1-p_j\right),
\end{align*}
where $w_i$ and $p_i$ are the edge weight and failur probability of the $i^{th}$ edge, $(v_i,v_{i+1})$, in the chain, for $i=1,\dots,k$.
\end{lemma}

In other words, the discounted weight of a chain can be expressed as the sum of the ``discounted weights'' of each \emph{edge} in the chain, i.e. $u(\kappa) = \sum_{i=1}^{k} w'_i,$
where $w'_i\equiv w_i \prod_{j=1}^{i} (1-p_j)$, where we refer to $\prod_{j=1}^{i} (1-p_j)$ as the \emph{discount factor}. 
 
The objective of Optimization (\ref{exact:exist1}) uses Lemma \ref{utility:chain} to express the total discounted weight of all matched cycles and chains, assuming non-uniform edge failure probabilities.
This is achieved using two sets of variables, $o_{ek}$ (the discount factor of edge $e$ at position $k$ in a chain) and $v_c$ (the success probability of cycle $c$).
Optimization (\ref{exact:exist1}) uses the following parameters:
\begin{itemize}
    \item $G$: kidney exchange graph, consisting of edges $e \in E$ and vertices $v \in V = P \cup N$, including patient-donor pairs $P$ and non-directed donors (NDDs) $N$
    \item $C$: a set of cycles on exchange graph $G$
    \item $L$: chain cap (max. number of edges in a chain)
    \item $w_e$: edge weights for each edge $e\in E$
    \item $w_c$: cycle weights for each cycle $c \in C$, defined as $w_c = \sum_{e\in c}w_e$
    \item $\delta^-(i)$: the set of edges into vertex $i$
    \item $\delta^+(i)$: the set of edges out of vertex $i$
    \item $p_e$: failure probability for edge $e\in E$
    
    
\end{itemize}

Edges between an NDD $n \in N$ and a patient-donor vertex $d \in P$ may only take position $1$ in a chain, while edges between two patient-donor pairs may take any position $ 2, \dots , L$ in a chain. For convenience, we define the function $\mathcal{K}$ for each edge $e$, such that $\mathcal{K}(e)$ is the set of all possible positions that edge $e$ may take in a chain.
\begin{align}
    \K(e) = 
    \begin{cases}
        \{1\}, \quad & e \text{ begins in } n\in N,\\
        \{2,\dots,L\}, \quad & e \text{ begins in } d \in P.
    \end{cases}
\end{align}
The following decision variables are used.
\begin{itemize}
    \item $z_c \in \{0,1\}$: $1$ if cycle $c$ is used in the matching, and $0$ otherwise
    \item $y_{ek} \in \{0,1\}$: $1$ if edge $e$ is used at position $k$ in a chain, and $0$ otherwise
    \item $o_{ek} \in [0,1]$: discount factor of edge $e$ at position $k$ in a chain
\end{itemize}

Our formulation is given in (\ref{exact:exist1}).
\begin{subequations}\label{exact:exist1}
    \begin{align}
     \max_{\y,\z,\o} \quad  & \sum_{e\in E} \sum_{k \in \K(e) } w_e y_{ek} o_{ek} + \sum_{c\in C}w_c z_c v_c \\
     s.t.\quad & \{\y,\z\} \in \mathcal{X},\\
    &  \sum_{\mathclap{{\tiny \begin{array}{c}e\in \delta^-(i)\wedge \\ k\in\K(e)\end{array}}}}  o_{ek}y_{ek} \geqslant \sum_{e\in \delta^+(i)} \frac{o_{e,k+1}y_{e,k+1}}{1-p_e} , &\nonumber \\
    &i\in P,k\in  \{1,\dots,L-1\}, \label{exp:constraint}\\
    & 0 \leqslant o_{ek} \leqslant 1-p_e, e\in E, k\in \K(e),\label{exp:more}\\
    & v_c = \prod_{e \in c} 1-p_e, c\in C, \label{exp:more1}
    \end{align}
 \end{subequations}
 where $\mathcal X$ denotes the set of feasible decision variables for the PICEF formulation of kidney exchange~\citep{Dickerson16:Position}, defined as
 {\small
 \arraycolsep=3pt\def\arraystretch{1.2}
 \begin{equation}
    \mathcal{X} = \left\{
    \begin{array}{ll} 
    \quad \sum\limits_{\mathclap{e\in \delta^-(i)}} \quad \quad \sum\limits_{\mathclap{k\in\K(e)}} y_{ek} + \sum\limits_{\mathclap{{\tiny \begin{array}{c} c\in C: \\ i\in c\end{array}}}} z_c \leqslant 1, & i\in P;\\
    \quad \sum\limits_{\mathclap{{\tiny \begin{array}{c} e\in \delta^-(i)\land\\ k\in\K(e) \end{array}}}}  y_{ek} \geqslant \quad \sum\limits_{\mathclap{e\in \delta^+(i)}} y_{e,k+1}, &{\arraycolsep=0pt \begin{array}{l}i\in P \\k\in \{1,...,L-1\}; \end{array}}\\
    \sum\limits_{e\in \delta^+(i)} y_{e1} \leqslant 1, & i \in N; \\
      y_{ek} \in \{0,1\}, & e\in E, k\in \K(e);\\
     z_c \in \{0,1\}, & c\in C;
     \end{array} \right. \label{exp:constraint1}
 \end{equation}

 }
 
 %
 %
The constraints of $\mathcal{X}$ are interpreted as follows: 1) the first constraint in (\ref{exp:constraint1}) requires that each patient-donor vertex $i\in P$ may only participate in one cycle or one chain; 2) the second constraint requires that each patient-donor vertex $i\in P$ can only have an outgoing edge at position $k + 1$ in a chain if it has an incoming
edge at position $k$; 3) the third constraint requires that each NDD $i \in N$ may only participate in one chain.

Constraints (\ref{exp:constraint}), (\ref{exp:more}), and (\ref{exp:more1}) define the discounted weight of chains and cycles.
We briefly describe how the discounted weight of cycles and chains are represented in this formulation:
 \begin{itemize}
 
     \item For a \emph{cycle}, the success probability is ${v_c = \prod_{e \in c} 1-p_e}$. Thus the discounted weight of all cycles is expressed as $\sum_{c\in C}w_c z_c v_c$.
     
     \item For a \emph{chain}, the discounted weight is expressed using Lemma \ref{utility:chain}. Consider the following example: suppose a $k$-chain consists of edges $e_1,\dots,e_k$. Suppose that $i$ is the \emph{first} patient-donor pair in this chain-- so $e_1$ is the edge \emph{into} $i$, and $e_2$ is the edge \emph{out of} $i$; that is, $e_1\in \delta^-(i)$ and $e_2 \in \delta^+(i)$. From constraints (\ref{exp:constraint})
     we have $o_{e_1,1} \geqslant \frac{1}{1-p_{e_2}}o_{e_2,2}$ for vertex $i$. The sums in constraint (\ref{exp:constraint}) contain no other terms, because $\mathcal{X}$ requires that only one edge into vertex $i$ and one edge out of vertex $i$ can be matched. Therefore, $(1-p_{e_2}) o_{e_1,1} \geqslant o_{e_2,2}$. 
     
     Similarly, $(1-p_{e_{j+1}}) o_{e_j,j} \geqslant o_{e_{j+1},j+1}$ for $j=2,\dots,k-1$. Since Optimization (\ref{exact:exist1}) is a maximization problem, the optimal values of variables $o_{e,j}$ will satisfy $o_{e_j,j} = \prod_{i=1}^j (1-p_{e_i})$, for $1\leqslant j \leqslant k$. Accordingly, $\sum_{e\in E} \sum_{k \in \K(e) } w_e y_{ek} o_{ek}$ represents the total discounted weight of all chains according to Lemma \ref{utility:chain}.

 \end{itemize}

 \subsection{MIP Reformulation of Optimization (\ref{exact:exist1})}\label{sec:main:tractability}
 
Although Optimization (\ref{exact:exist1}) exactly maximizes expected edge weight under non-identical edge failure probabilities, it is a nonconvex optimization problem. In this section, we reformulate it as a mixed-integer linear program which can be solved usings general-purpose solvers. Proposition \ref{linear2} concludes our results; the main idea is to define a set of new variables $O_{ek}$ to replace $y_{ek}o_{ek}$ in Optimization (\ref{exact:exist1}).

\begin{proposition}\label{linear2}
Optimization (\ref{exact:exist1}) is equivalent to
{\small
 \begin{equation}\label{linear:closed2}
     \begin{aligned}
     \max_{\y,\z,\O,\o} \quad  & \sum_{e\in E} \sum_{k \in \K(e) } w_e  O_{ek} + \sum_{c\in C}w_c z_c ( \prod_{e \in c} 1-p_e )\\
     s.t.\quad & \{\y,\z\} \in \mathcal{X},\\
     \quad & \{\y,\O,\o\} \in \mathcal{X'},
     \end{aligned}
 \end{equation}
 where $\mathcal{X}$ follows the definition in (\ref{exact:exist1}), and $\mathcal{X'}$ is defined as
 {\small
 \begin{equation}
    \begin{aligned}
    \mathcal{X'} = \left\{
    \begin{array}{ll}
         \displaystyle \sum_{e\in \delta^-(i) \land k\in\K(e)} O_{ek} \geq \sum_{e\in \delta^+(i)} \frac{O_{e,k+1}}{1-p_e};\\
         i\in P,k\in \{1,\dots,L-1\},\\ O_{ek} \leq y_{ek},  e\in E,  k\in \K(e);\\
         O_{ek} \leq o_{ek},  e\in E ,k \in  \K(e);\\
         O_{ek}\in [0,1], e\in E, k\in \K(e);\\
         0 \leq o_{ek} \leq 1-p_e, e\in E, k\in \K(e).
    \end{array} \right\}
    \end{aligned}
\end{equation}
}
}
 
\end{proposition}
Appendix~\ref{app:example} gives an explicit example of how to use the reformulation in Proposition~\ref{linear2} to model a kidney exchange graph.  Optimization (\ref{linear:closed2}) can be solved using standard solvers such as CPLEX and Gurobi.

\subsubsection*{Scalability}
We compare our model size with state-of-the-art approaches in literature. We summarize all approaches in Table \ref{tab:method}. The size of each model (the number of variables and constraints) is expressible in terms of the chain cap $L$, and the number of edges ($|E|$), cycles ($|C|$), total vertices ($|V|$), NDD vertices ($|N|$), and patient-donor pair vertices $|P|$. For ease of exposition we assume $|N| = O(|V|)$ and $|P| = O(|V|)$.

Our size is comparable with PICEF, while accounting for non-identical failure probabilities.
DPS-18~\citep{Dickerson18:Failure} considers non-identical failure probabilities at the cost of representing every single chain and cycle as a decision variable, and thus this model grows exponentially with the chain cap $L$; in contrast, the number of variables in our formulation is polynomial in $L$. Real exchanges often use a cycle cap of $3$, which is sufficiently small that all cycles can be enumerated in practice--even on realistic graphs with hundreds of vertices. If exchanges grow much larger in the future (e.g., thousands of vertices), or if cycle lengths are increased substantially, we further propose a branch-and-price implementation to solve the corresponding problems brought by huge $|C|$ in Appendix \ref{app:branchAndPrice}.


\section{EXTENSIONS TO MEAN-RISK KIDNEY EXCHANGE MODEL}

Next we introduce a kidney exchange model which balances both the \emph{mean expected weight} and the \emph{worst-case} weight (``risk'') of a matching, using known non-identical edge failure probabilities. We achieve this balance using a conditional value-at-risk (CVaR) objective. This approach is motivated by the fact that the \emph{expected} weight of a matching can be misleading when the \emph{worst-case} outcome can be arbitrarily bad (see \S~\ref{sec:example}). This is especially true in kidney exchange, where a single edge failure can impact an entire cycle or chain. 




\subsection{Mean-risk Model}\label{sec:mean-risk}


At a high level, the CVaR objective for kidney exchange is expressed as
$$\displaystyle \mu + \gamma \times \mu_{\alpha},$$ 
where $\mu$ is the expected matching weight and $\mu_\alpha$ is the $\alpha \times 100\%$ ($\alpha \in (0,1]$) \emph{worst-case} mean weight--that is, the mean matching weight in the worst $\alpha\times 100\%$ of all outcomes. The parameter $\gamma$ is set by the user, and controls the trade-off between average performance and the \emph{risk} of the solution. 

For tractability and simplicity, we define $\mathbf{W}$ as an $|E|$-dimensional vector with 
$$ W_e =  -\sum_{k \in \K(e) } y_{ek}  -\sum_{c\in C} \1(e\in c)z_c ,\quad \forall e\in E.$$
That is, $W_e=-1$ if edge $e$ is used, and $W_e=0$ otherwise. We use $\w \in \mathbb{R}^{|E|}$ to represent the random discounted edge weights under known edge failure probabilities. Correspondingly, $\langle \w,\W \rangle$ represents the loss (negative weight) of a matching.
The $\alpha \times 100\%$ worst-case (highest) mean loss is equivalent to the CVaR objective~\citep{rockafellar2000optimization} at level $\alpha$. 
The corresponding optimization problem is expressed in (\ref{kidney:c:closed}), by introducing an auxiliary variable $d$. We use $(x)^+$ to denote $\max(0,x)$, and the expectation in (\ref{kidney:c:closed}) is taken over the distribution of random edge weights under the known edge failure probabilities. As before, $\mathcal X$ denotes the set of feasible matchings using the PICEF formulation.

{\small
\begin{equation}\label{kidney:c:closed}
    \begin{aligned}
        \min_{\y,\z,d} \quad  &\E( \langle \w,\W \rangle) + \gamma \left[d+\frac{1}{\alpha}\E \left[ (\langle \w,\W \rangle - d)^+\right] \right]\\
        s.t.\quad & \{\y,\z\} \in \mathcal{X}.
    \end{aligned}
\end{equation}
}

\subsection{An SAA-based Approach for Optimization (\ref{kidney:c:closed})}\label{sec:mean-risk:optimization}

The main difficulty in solving Optimization (\ref{kidney:c:closed}) is that term $ \displaystyle \E \left[ (\langle \w,\W \rangle - d)^+\right]$
does not have a simple closed-form reformulation under the known edge failure probabilities. Thus, we propose an approach based on Sample Average Approximation (SAA)~\citep{Anderson15:Finding} to solve (\ref{kidney:c:closed}). 
The main idea is to first sample $N$ realizations of edge existence according to the known edge failure probabilities; for each realization we formulate a mixed-integer linear program representing the matching weight under this realization. Finally, we combine all $N$ models to obtain an optimization problem that is (approximately) equivalent to Optimization (\ref{kidney:c:closed}) based on these $N$ realizations.
%
%
Algorithm~\ref{alg_saa} gives a pseudocode description of this approach.

\begin{algorithm}
\vspace{-0mm}
\footnotesize
\caption{SAA}\label{alg_saa}
\begin{algorithmic}[1]
\STATE \textbf{Initialization: } $N $;
\STATE \textbf{STEP 1}:
\STATE Sample $N$ edge existence realizations $\{\hat{r}_{en} \in\{0,1\}, \forall e \in E\}$, $n=1, \dots, N$;
\STATE \textbf{STEP 2}:
\STATE Solve Optimization (\ref{SAA:existence}).
\end{algorithmic}
\end{algorithm}

This algorithm has only two steps: first it samples $N$ \emph{realizations} of edge existence from the known edge failure probabilities, where $\hat{r}_{en}$ is $1$ if edge $e$ succeeds in realization $n$, and $0$ if it fails.
These realization variables are used as input to Optimization (\ref{SAA:existence}), which uses decision variables $\hat{\W}_n$ to represent the \emph{realized} edge discount factor for realization $n$--that is, $\hat{W}_{en}$ is $1$ if edge $e$ is matched and succeeds in realization $n$ and $0$ otherwise (see Appendix~\ref{app:equivalence} for  details).
Using these decision variables, the objective of Optimization (\ref{SAA:existence}) includes two terms: the mean matching weight, and the CVaR objective--both approximated using all $N$ samples (i.e., the sample-average approximation). 
Thus, Optimization (\ref{SAA:existence}) represents the SAA of (\ref{kidney:c:closed}) under the $N$ sampled realizations. 
\begin{proposition}\label{prop:cvar}
Optimization (\ref{SAA:existence}) is equivalent to the SAA of (\ref{kidney:c:closed}) under $N$ edge existence realizations represented by $\hat{r}_{en}$, with
 \begin{equation}\label{SAA:existence}
     \begin{aligned}
     \min \quad  &\frac{1}{N}\sum_{n=1}^N \langle \w, \hat{\W}_n \rangle  + \gamma \left(d+\frac{1}{\alpha}\frac{1}{N}\sum_{n=1}^N \Pi_n \right) \\
     s.t.\quad & \hat{W}_{en} =  -\sum_{\mathclap{k \in \K(e) }} O_{ekn}  -\sum_{c\in C} \1(e\in c)z_c v_{cn} ,\forall e,n,\\
     & \Pi_n \geqslant 0, \forall n, \\
     & \Pi_n \geqslant \langle \w,\hat{\W}_n \rangle - d, \forall n,\\
     & \{\y,\z\} \in \mathcal{X},\\
     & \{\y,\z\} \in \mathcal{X'},\\
     & o_{ekn} \leqslant \hr_{en}, \forall e,k,n,\\
   & v_{cn} = \min_{e\in c}\{\hr_{en}\}, \forall c,n,
     \end{aligned}
 \end{equation}
  where $\mathcal{X}$ follows the definition in (\ref{exact:exist1}), and $\mathcal{X'}$ is defined as 
   \begin{equation*}
    \begin{aligned}
     \left\{
    \begin{array}{ll}
         \displaystyle \sum_{e\in \delta^-(i) \wedge k\in\K(e)} O_{ekn} \geqslant \sum_{e\in \delta^+(i)}O_{e,k+1,n},\\
         \forall i\in P,k\in \{1,\dots,L-1\}, n\in\mathcal{N}; \\
        O_{ekn} \leqslant y_{ek},  e\in E, k\in \K(e), n \in \mathcal{N};\\
        O_{ekn} \leqslant o_{ekn},  e\in E, k\in \K(e), n \in \mathcal{N};\\
        o_{ekn}, O_{ekn}\in [0,1],   e\in E, k\in \K(e), n \in \mathcal{N};\\
         \mathcal{N}= \{1, \dots, N\}.
    \end{array} \right\}
    \end{aligned}
\end{equation*}
  
\end{proposition}
 
Optimization (\ref{SAA:existence}) can be understood by viewing $\langle \w, \hat{\W}_n \rangle$ as the realized edge weight under the $n$-th realization with matching $\{\y,\z\}$. 

\section{COMPUTATIONAL EXPERIMENTS}

%
First, we benchmark our tractable model for non-identical edge failure probabilities~(\ref{linear:closed2}) (``KEP-NP'') against previous approaches, with the \emph{stochastic} (i.e., max-expected-weight) objective. We find that our approach  outperforms two leading previous methods: PICEF without edge failure probabilities~\citep{Dickerson16:Position} (``KEP''), and PICEF with identical edge failure probability~\citep{Dickerson18:Failure} (``KEP-IP''). Second, we compare our CVaR model (\ref{SAA:existence}) (``CVAR'') against KEP, KEP-IP, and KEP-NP; to our knowledge, there are no other tractable approaches using the CVaR objective in our setting. Finally, we briefly present the running time of all implemented approaches.

\subsection{Stochastic Objective}\label{sec:exp1}


We use two sets of $32$ randomly-generated graphs, one with $64$ nodes each and one with $128$ nodes each. These graphs resemble the structure of real exchanges, and are generated using anonymized data from the United Network for Organ Sharing (UNOS) US-wide kidney exchange.\footnote{See \url{https://optn.transplant.hrsa.gov/data} for more information on UNOS.} We simulate edge existence uncertainty by randomly assigning each edge in each graph a failure probability, independently uniformly distributed on $[0.1, 0.9]$; for simplicity, we set all edge weights to 1. 
We use cycles of length 2 and 3, and chains up to length 4--which are the standard limits in fielded exchanges (including UNOS).
For KEP-IP, we assume $p_e=0.5$ for all edges (the correct mean edge failure probability). For each random exchange graph we first find the optimal matching according to each approach (KEP, KEP-IP, and KEP-NP). We then generate 200 \emph{realizations} of the exchange graph, according to each edge's (randomly generated) failure probability. We then calculate the \emph{realized} weight of the optimal matching for each method, accounting for failed edges (cycles with any failed edges receive zero weight, and chains only receive weight for consecutive successful edges, beginning with the first). We also calculate the \emph{omniscient} matching for each realization, i.e., the maximum matching weight \textit{after} observing edge failures.

\paragraph{Metric: Percentage of Omniscient Weight.}  We compare all approaches against the omniscient matching weight, which is a strict upper bound on performance for any matching approach. Let $\WOPT{}$ be the omniscient-optimal matching weight for a particular exchange, and a particular realization; let $W_M$ be the realized matching weight for a non-omniscient method. We calculate the percentage of $\WOPT{}$ achieved by each matching method, for a particular realization, as
$\displaystyle \% \text{OPT} \equiv 100 \times W_M / \WOPT{}.$ Figure~\ref{fig:results} (left column) shows $\displaystyle \% \text{OPT}$ for all exchange graphs, over all 200 realizations, for $64$-node graphs (top) and $128$-node graphs (bottom). Our method (KEP-NP) improves expected matching weight compared to previous methods KEP and KEP-IP.

\subsection{CVaR Objective}

We implement CVAR (\S~\ref{sec:exp1}) using $N=10$ simulated edge realizations, with $\gamma=10$, and $\alpha=0.5$.

\begin{figure*}[ht]
\centering
\begin{subfigure}[b]{0.27\textwidth}
        \centering\includegraphics[width=\linewidth]{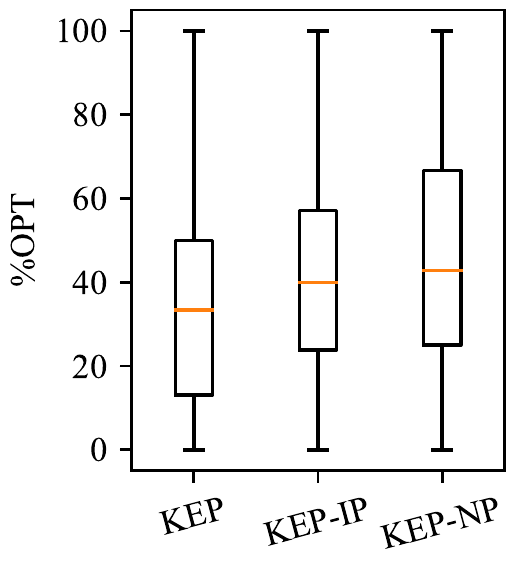}
    \caption{$\% \text{OPT}$ for $64$-node graphs.\label{fig:pct-opt-64}}
        \end{subfigure}%
    \hfill
    \begin{subfigure}[b]{0.27\textwidth}
        \centering\includegraphics[width=\linewidth]{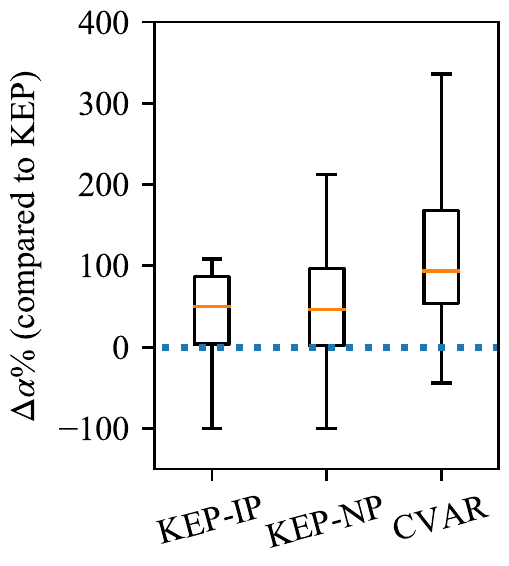}
    \caption{$\Delta \alpha\%$ for $64$-node graphs.\label{fig:alpha-64}}
        \end{subfigure}%
    \hfill
    \begin{subfigure}[b]{0.35\textwidth}
    \centering\includegraphics[width=\linewidth]{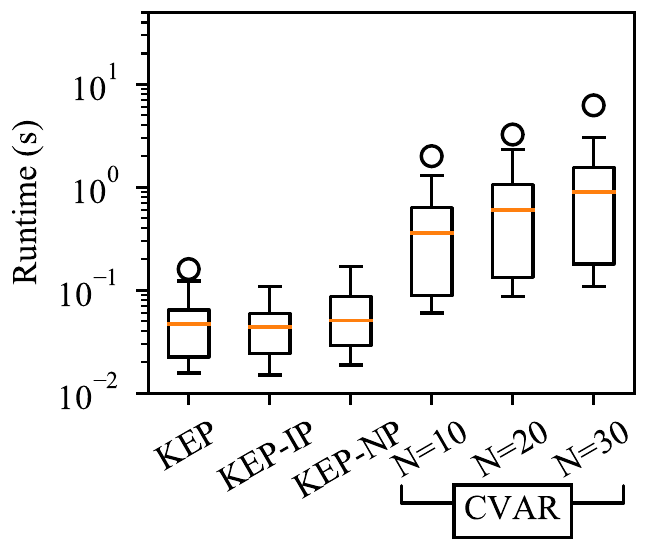}
\caption{Timing for $64$-node graphs.\label{fig:timing-64}}
    \end{subfigure}%
    
 \begin{subfigure}[b]{0.27\textwidth}
        \centering\includegraphics[width=\linewidth]{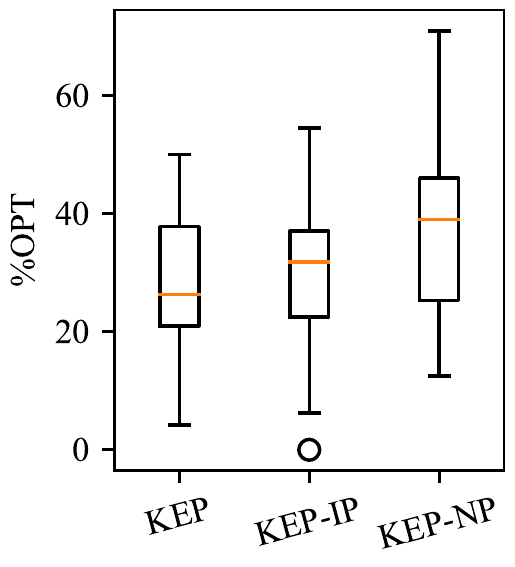}
    \caption{$\% \text{OPT}$ for $128$-node graphs.\label{fig:pct-opt-128}}
        \end{subfigure}%
    \hfill
    \begin{subfigure}[b]{0.27\textwidth}
        \centering\includegraphics[width=\linewidth]{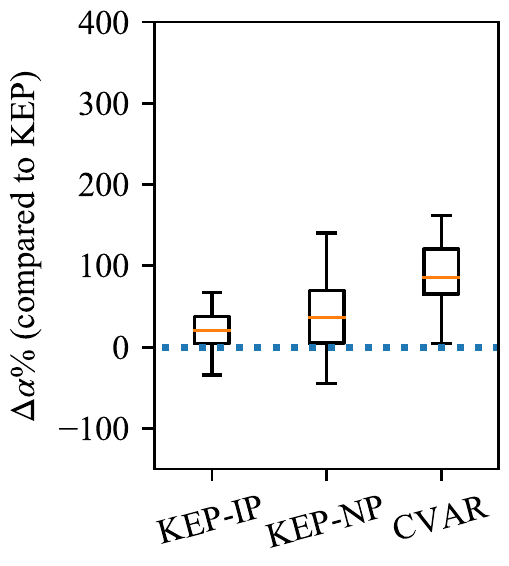}
    \caption{$\Delta \alpha\%$ for $128$-node graphs.\label{fig:alpha-128}}
        \end{subfigure}%
    \hfill
    \begin{subfigure}[b]{0.35\textwidth}
    \centering\includegraphics[width=\linewidth]{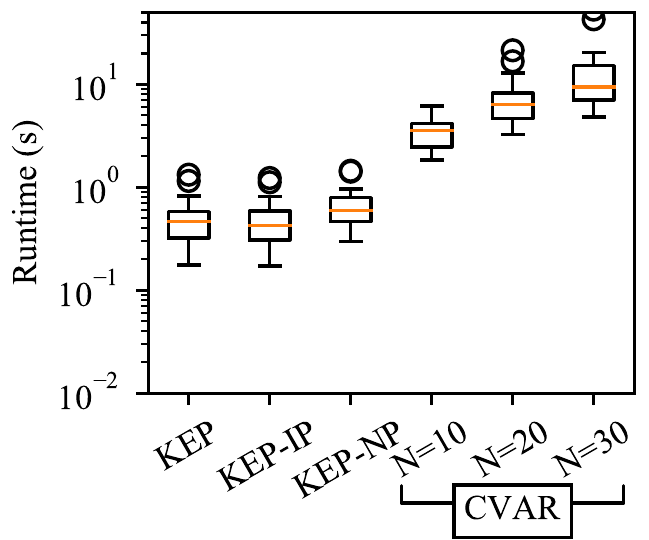}
\caption{Timing for $128$-node graphs.\label{fig:timing-128}}
    \end{subfigure}%
    \caption{Boxplots of $\% \text{OPT}$ (left column), $\Delta \alpha\%$ (center column), and timing (right column) for each matching approach, over $32$ random graphs with $64$ nodes (top row) and $128$ nodes (bottom row). The horizontal line at the center of each box plot indicates the median; the upper and lower edges of the box indicate the first and third quartiles; the whiskers extend 1.5 times the interquartile range beyond quartile 1 and 3.}
    \label{fig:results}
\end{figure*}

\paragraph{Metric: $\alpha$\% Worst-Case Mean.} CVAR is designed to maximize the $\alpha$\% worst-case mean matching weight; thus, we use this metric for each matching approach. For each graph, and each matching approach, we calculate the mean of the $\alpha$\% lowest realized matching weights (over all 200 realizations). We compare each method to KEP, which assumes $p_e=0$. Let $\mu^{\alpha}_{P}$ be the $\alpha$\% lowest-realized matching weights for KEP, and let $\mu^\alpha_M$ be the same, for a different matching approach; we calculate a ratio as follows:
$\displaystyle \Delta \alpha\% \equiv 100 \times (\mu^\alpha_M - \mu^\alpha_P )/ \mu^\alpha_P$ .
Figure~\ref{fig:results} (middle column) shows $\displaystyle \Delta \alpha\%$ for all 32 exchange graphs. CVAR clearly improves the $\alpha\%$ worst-case mean matching weight, over other methods (including our new formulation for inhomogeneous edge failure probabilities, KEP-NP). 

\paragraph{Timing}
Figure~\ref{fig:results} (right column) shows solver time required for each method. Our new formulation (KEP-NP), requires nearly the same runtime as KEP (a deterministic PICEF model). As expected CVAR requires more time--and it increases with the number of samples ($N$).

%
\section{CONCLUSION}

In fielded kidney exchanges, planned transplants \emph{fail} for a variety of reasons. 
Due to the cycle- and chain-like swaps used by exchanges, a single failed transplant can ``cascade'' throuh an exchange, causing several other transplants to fail.
These failures are common (UNOS estimates that about $85\%$ of its planned transplants fail~\citep{Leishman19:Challenges}); failures cause patients to face longer waiting times, and incur the additional costs and burden of dialysis.

We consider a setting where the failure probability of each potential transplant (\emph{edge}) is known, and the \emph{kidney exchange clearing problem} is to select a set of transplants that maximize a mathematical objective subject to this uncertainty.
The choice of objective is important, particularly in kidney exchange: a \emph{deterministic} approach (which ignores potential failures) may na\"ively select long cycles or chains, which have high likelihood of failure.
On the other hand, a \emph{robust} approach (which protects against the worst-case outcome) is often too conservative, because in kidney exchange, the worst-case outcome is often that \emph{all transplants} fail.
We consider two objectives: maximizing the \emph{expected} weight, and maximizing the conditional value-at-risk (CVaR).
We are not the first to investigate these objectives in this setting.
However, state-of-the-art approaches either assume that all edges have identical failure probabilities, or their algorithms scale exponentially in the size of the input---and are intractable for realistic exchanges.

We propose the first \emph{scalable} approaches for kidney exchange with non-identical edge failure probabilities, for both the stochastic and CVaR objectives.
For the max-expected weight objective our approach is exact, and clearly outperforms prior approaches that assume identical edge failure probabilities---with marginally longer runtime.
For the CVaR objective we use a sample-average-approximation-based method, which outperforms comparable state-of-the-art approaches, even with a small number of samples.
We formulate both of our approaches as mixed integer linear programs, which are solvable with off-the-shelf commercial solvers such as CPLEX or Gurobi.

There are several areas for future work. 
Our model assumes perfect knowledge of edge failure probabilities--while in reality only rough estimates of these probabilities are available.
Furthermore, slight over- or under-estimation of these probabilities can impact the matching weight~\citep{Dickerson18:Failure}---something we did not address in this work.

We emphasize that the choice of objective is important in kidney exchange, as different objectives (or a different weighting of multiple objectives) can drastically change the outcome. 
Before implementing any of these approaches, it is necessary to understand the priorities of the relevant stakeholders, their appetite for risk, and whether these priorities align with our mathematical objectives~\citep{Noothigattu18:Making,Freedman20:Adapting}.

Finally, each of the approaches discussed in this paper may negatively impact some exchange participants.
For example, \emph{highly-sensitized} patients are often sicker and harder to match than other patients; transplants involving highly-sensitized patients are thus often riskier than other transplants.
A risk-averse or stochastic objective function would likely de-prioritize highly-sensitized patients, ignoring them for lower-risk matches.
Thus, new objective functions and other modeling choices will likely raise concerns of \emph{fairness} for different patients, or groups of patients, within an exchange~\citep{Roth05:Pairwise,Yilmaz11:Kidney,McElfresh18:fair}.
%


\noindent\textbf{Acknowledgments.}
Dickerson and McElfresh were supported in part by NSF CAREER IIS-1846237, DARPA GARD \#HR112020007, DARPA SI3-CMD \#S4761, DoD WHS \#HQ3420F0035, NIH R01 NLM-013039-01, and a Google Faculty Research Award.


 \bibliographystyle{plainnat}  
 \bibliography{refs}

%
%
%
%

\clearpage
\appendix

\section{Proof of Lemma \ref{utility:chain}}\label{app:lem1}
\begin{proof}
The expected discounted weight of a chain with $k$ edges is expressed as
 \begin{align*}
    u(k) =& \sum_{i=2}^{k}p_i (\sum_{j=1}^{i-1}w_j)\prod_{j=1}^{i-1} (1-p_j)\nonumber \\
     &+ (\sum_{i=1}^{k}w_i) \prod_{i=1}^{k} (1-p_i).\label{lem:former}
 \end{align*}
%

%
The coefficient on weight $w_i$ (the $i^{th}$ edge in the chain), for any $1 \leq i\leq k$, is expressed as $\prod_{j=1}^{i}(1-p_j)$. 
Thus, 
$$ u(k) = \sum_{i=1}^k w_i\prod_{j=1}^{i} (1-p_j).$$

\end{proof}

\section{Operationalizing Proposition~\ref{linear2}}\label{app:example}
In this section, we give an explicit instantiation of the mixed-integer linear program of Proposition~\ref{linear2}.  We provide an example here for the compatibility graph shown in Figure~\ref{fig:example-graph}, where there are two cycles, five edges, and several feasible chains. Suppose cycle $1$ contains Edge $1$ and $2$, and cycle $2$ contains Edge $3$ and $4$. Additionally, the chain capacity is $L = 2$ in Figure \ref{fig:example-graph}.
{\small
 \begin{equation}\label{example:fig:1}
     \begin{aligned}
     \max_{\y,\z,\O,\o} \quad  & \sum_{e=1}^4 w_e  O_{e2} + w_5O_{51} \\
     & \quad + (w_1+w_2)z_1(1-p_1)(1-p_2) \\
     & \quad + (w_3+w_4)z_2(1-p_3)(1-p_4)\\
     s.t.\quad & y_{51}+y_{22}+y_{42} + z_1+z_2 \leqslant 1,\\
     &  y_{12} + z_1 \leqslant 1,\\
     &  y_{32} + z_2 \leqslant 1,\\
    & y_{51} \geqslant y_{12}+y_{32},\\
    & y_{51} \leqslant 1, \\
    &  y_{e2} \in \{0,1\},  e\in \{1,\dots,4\},\\
    & y_{51} \in \{0,1\},\\
    & z_c \in \{0,1\},  c \in \{1,2\},\\
    & O_{51} \geqslant \frac{O_{12}}{1-p_1} + \frac{O_{32}}{1-p_3},\\
    & O_{e2} \leq y_{e2}, e\in \{1,\dots,4\},\\
    & O_{51} \leq y_{51},\\
    & O_{e2} \leq o_{e2},  e\in \{1,\dots,4\}, \\
    & O_{51} \leq o_{51},  \\
    & O_{e2}\in [0,1], e\in \{1,\dots,4\},\\
    & O_{51}\in [0,1],\\
    & 0 \leq o_{e2} \leq 1-p_e, e\in \{1,\dots,4\},\\
    & 0 \leq o_{51} \leq 1-p_5.
     \end{aligned}
 \end{equation}
}


\section{Optimization (\ref{kidney:c:closed}) under one realization of edge existence}\label{app:equivalence}

Before showing the equivalence between the SAA of (\ref{kidney:c:closed}) and (\ref{SAA:existence}), we first obtain the objective value of (\ref{kidney:c:closed}) under one fixed realization of edge existence.
The objective value of (\ref{kidney:c:closed}) is obtained in (\ref{exist:picef}), where we assume the fixed realization is $r_e \in \{0,1\}$, $e\in E$, where $1$ means the edge exists, and $0$ otherwise.

In (\ref{exist:picef}), we use two sets of variables $o_{ek} \in \{0,1\}$ and $v_c \in \{0,1\}$, which indicate the validity of chains and cycles, respectively. 
 \begin{itemize}
     \item  For any cycle $c$, $c$ is only valid ($v_c = 1$) if all edges in $c$ exist. Therefore, we restrict $v_c = \min_{e\in c}\{r_{e}\}$.
     \item For any chain, an edge $e$ at position $k$ is only valid ($o_{ek} =1$) if 1) this edge exists ($r_e = 1$) and 2) the prior edges in this chain are valid too. Therefore, we use $o_{ek} \leqslant r_{e}$ to guarantee this edge $e$ exists. The constraint (\ref{app:B:constraint}) serves the goal to guarantee the prior edges are valid. To see this point, we consider the following example. Suppose edge $e_1 \in \delta^-(i)$ and $e_2 \in \delta^+(i)$. Both edges are selected in one chain with $y_{e_1,k}=1$, $y_{e_2,k+1} = 1$. If edge $e$ fails ($r_{e_1}=0)$, then $o_{e_1,k}=0$ restricting $o_{e_2,k+1}$ to be zero too. Therefore, all the edges after position $k$ in this chain will be invalid.
 \end{itemize}

{\bf }

{\small
\begin{subequations}\label{exist:picef}
         \begin{align}
     \min_{\y,\z,\o,\v,d} \quad  &\left( -\sum_{e\in E} \sum_{k \in \K(e) } w_e y_{ek} o_{ek} - \sum_{c\in C}w_c z_c v_c\right) \nonumber\\
     &+ \gamma \Bigg[d+\frac{1}{\alpha}(-\sum_{e\in E} \sum_{k \in \K(e) } w_e y_{ek} o_{ek} \nonumber \\
     &- \sum_{c\in C}w_c z_c v_c - d)^+ \Bigg] \\
     s.t.\quad & \{\y,\z\} \in \mathcal{X},\\
    & \sum_{{\tiny \begin{array}{c} e\in \delta^-(i) \land \\ k\in\K(e) \end{array}}} o_{ek}y_{ek} \geq \sum_{e\in \delta^+(i)}o_{e,k+1}y_{e,k+1}, \nonumber \\ 
    & i\in P,k\in \{1,\dots,L-1\}, \label{app:B:constraint}\\
    & o_{ek} \leqslant r_{e},  e\in E,k \in \K(e),\\
     & v_c = \min_{e\in c}\{r_{e}\},  c \in C; \\
    & o_{ek} \in [0,1],  e\in E, k\in \K(e).
     \end{align}
 \end{subequations}
 }
Optimization (\ref{exist:picef}) has a tractable reformulation as shown in Proposition \ref{linear}.
\begin{proposition}\label{linear}
Optimization (\ref{exist:picef}) is equivalent to
{\small
 \begin{equation}\label{linear:closed}
     \begin{aligned}
     \min_{\y,\z,\o,\v,d} \quad  &\langle \w, \hat{\W} \rangle + \gamma \left[d+\frac{1}{\alpha}(\langle \w, \hat{\W} \rangle - d)^+ \right] \\
     s.t.\quad & \{\y,\z\} \in \mathcal{X}, \mathcal{X'}\\
     & \hat{W}_{e} =  -\sum_{k \in \K(e) } O_{e,k}  -\sum_{c\in C} \1(e\in c)z_{c} v_c ,\forall e,\\
     & o_{e,k} \leqslant r_{e}, \forall e,k,\\
   & v_c = \min_{e\in c}\{r_{e}\}, \forall c.
     \end{aligned}
 \end{equation}
 where $\mathcal{X'}$ is defined as
 \begin{equation*}
     \begin{aligned}
 \mathcal{X'} = \left\{
    \begin{array}{ll}
      \sum_{e\in \delta^-(i) \wedge k\in\K(e)} O_{e,k} \geqslant \sum_{e\in \delta^+(i)}O_{e,k+1}, \\
      i\in P,k\in \{1,\dots,L-1\};\\
      O_{e,k} \leqslant y_{e,k},  e\in E, k \in \K(e);\\
     O_{e,k} \leqslant o_{e,k},  e\in E,k \in \K(e);\\
      o_{e,k}, O_{e,k}\in [0,1], e\in E, k\in \K(e).
 \end{array}
 \right\}
  \end{aligned}
 \end{equation*}
 }
\end{proposition}
By comparing Optimization (\ref{linear:closed}) and (\ref{SAA:existence}), it is easy to see that the objective value of (\ref{SAA:existence}) equals the average realized weights of $N$ realizations. Therefore, we get the conclusion that Optimization (\ref{SAA:existence}) is equivalent to the SAA of Optimization (\ref{kidney:c:closed}).

\section{A branch and price implementation}\label{app:branchAndPrice}
pIn this section, we present a method for scaling our model to graphs with high cycle capacities. Theoretically, the number of cycles of length at most $M$ is $O(|P|^M)$, making explicit representation and enumeration of all cycles infeasible for large enough instances. To solve this problem, we propose a branch and price algorithm, which uses column generation to incrementally consider the possible cycles in a graph. Similar ideas in other kidney exchange problems have also been explored \citep{Glorie14:Kidney,Dickerson16:Position}. We show that our formulation with non-identical failure probabilities also scales well with large cycle numbers.

The detailed procedure is introduced as follows; for convenience, we use a vector $\bm{X}$ to denote the solution $\bm{X} = [\y,\z]$. First, we define a set $\X_f$ that indicates the fixed components in the solution $\bm{X}$. For example, $\X_f=\{X_i = 0, X_j = 1\}$ means the $i$-th and $j$-th components in $\bm{X}$ are fixed to $0$ and $1$, respectively. Our algorithm begins with $\X_f = \emptyset$. Next, an LP relaxation (\ref{branchPrice:LP}) based on a (random) subset of cycles $C'$ ($C' \subset C$) is solved.

 \begin{subequations}\label{branchPrice:LP}
     \begin{align}
     \max_{\y,\z,\O,\o} \quad  & \sum_{e\in E} \sum_{k \in \K(e) } w_e  O_{ek} + \sum_{c\in C'}w_c z_c \left( \prod_{e \in c} 1-p_e \right)\\
     s.t.\quad &  \sum_{e\in \delta^-(i)} \sum_{ k\in\K(e)}  y_{ek} + \sum_{c\in C': i\in c} z_c \leqslant 1, i\in P, \label{branchPrice:cons}\\
    &\sum_{e\in \delta^-(i)\wedge k\in\K(e)}  y_{ek} \geqslant \sum_{e\in \delta^+(i)} y_{e,k+1},\\
    &i\in P, k\in \{1,\dots,L-1\},\\
     &\sum_{e\in \delta^+(i)} y_{e1} \leqslant 1, i \in N ,\\
      &y_{ek} \in [0,1], e\in E, k\in \K(e),\\
     &z_c \in [0,1], c\in C',\\
      &  \sum_{e\in \delta^-(i) \wedge k\in\K(e)} O_{ek} \geqslant \sum_{e\in \delta^+(i)} \frac{ O_{e,k+1}}{1-p_e},\\
      &i\in P,k\in \{1,\dots,L-1\},\\ &O_{ek} \leqslant y_{ek},  e\in E,  k\in \K(e),\\
       &  O_{ek} \leqslant o_{ek},  e\in E ,k \in  \K(e),\\
    &     O_{ek}\in [0,1], e\in E, k\in \K(e),\\
     &    0 \leqslant o_{ek} \leqslant 1-p_e, e\in E, k\in \K(e).
     \end{align}
 \end{subequations}
 
The following step is to find positive price cycles: cycles that have the potential to improve the objective value if included in the model. The price of a cycle $c$ is defined as $\left[w_c  \prod_{e \in c} (1-p_e) - \sum_{i\in c } \lambda_i \right]$, where $\lambda_i$ are the dual values corresponding to the constraints (\ref{branchPrice:cons}). While there exist any positive price cycles, optimality of the reduced LP has not yet been proved. This can be evidenced from Proposition \ref{prop:pricing}, which can be proved through the strong duality of linear programming.
\begin{proposition}\label{prop:pricing}
Suppose the dual variables corresponding to the constraints (\ref{branchPrice:cons}) are $\lambda_i$, $i \in P$. Then the optimal $\lambda_i$, $i\in P$, satisfy
\begin{align*}
    w_c  \prod_{e \in c} (1-p_e) - \sum_{i\in c } \lambda_i \leqslant 0.
\end{align*}
\end{proposition}

Therefore, we incrementally add (one or more) cycles that have positive prices, i.e. $ w_c  \prod_{e \in c} (1-p_e) - \sum_{i\in c } \lambda_i > 0$ into $C'$ until no positive price cycles exist in $C$. Afterwards, if the optimal solutions of the relaxed LP, i.e. (\ref{branchPrice:LP}), are integral, then they are the desired optimal solutions. Otherwise, branching occurs by following the standard branch-and-bound tree search. For example, suppose the $i$-th component of $\bm{X}$ is fractional, then we fix $X_i=0$ or $X_i = 1$. We record these fixed components in set $\X_f$ and repeat above procedures with the new set $\X_f$. We conclude the  above discussions in the following Algorithm \ref{alg:brach_price}. By running BranchAndPrice($G$, $\emptyset$), we obtain the optimal solution.
\begin{algorithm}
\small
\caption{BranchAndPrice($G$, $\X_f$)}\label{alg:brach_price}
\begin{algorithmic}[1]
\STATE Generate a subset $C' \subset C$;
\STATE Solve LP relaxation (\ref{branchPrice:LP}) based on $C'$ and fixed components in $\X_f$;
\WHILE{$\max_{c\in C}w_c  \prod_{e \in c} (1-p_e) - \sum_{i\in c } \lambda_i > 0 $}
\STATE {Add $c^* = \argmax_{c\in C}w_c  \prod_{e \in c} (1-p_e) - \sum_{i\in c }$ to $C'$;}
\ENDWHILE
\STATE $\bm{X} =[\y,\z]\leftarrow $ solve LP relaxation (\ref{branchPrice:LP}) based on $C'$ and fixed components in $\X_f$;
\IF{$\bm{X}$ is fractional}
\STATE {Find fractional binary variable $X_i \in \bm{X}$ closest to $0.5$;}
\STATE {$\bm{X}_1 = $ BranchAndPrice($G,\X_f\cup X_i=0$);}
\STATE {$\bm{X}_2 = $ BranchAndPrice($G,\X_f\cup X_i=1$);}
\STATE {Return $\bm{X}_1$ or $\bm{X}_2$ that gives larger objective values in the original Problem (\ref{linear:closed2}).}
\ELSE
\STATE{Return $\bm{X}$}.
\ENDIF
\end{algorithmic}
\end{algorithm}

\end{document}